\documentclass[10pt,twocolumn,letterpaper]{article}

\usepackage{cvpr}              

\usepackage{graphicx}
\usepackage{amsmath}
\usepackage{amssymb}
\usepackage{booktabs}
\usepackage{amsfonts}       
\usepackage{nicefrac}       
\usepackage{microtype}      
\usepackage{epsfig}
\usepackage{wrapfig}
\usepackage{amsmath}
\usepackage{amssymb}
\usepackage{graphicx}
\usepackage{enumitem}
\usepackage{xspace}
\usepackage{booktabs}
\usepackage{bm}
\usepackage{tabulary,multirow,overpic}
\usepackage{subcaption}
\newcommand{\tablestyle}[2]{\setlength{\tabcolsep}{#1}\renewcommand{\arraystretch}{#2}\centering\scriptsize}

\newlength\savedwidth

\usepackage[table]{xcolor}
\definecolor{Graylight}{gray}{0.9}

\newcommand{\Tref}[1]{Table~\ref{#1}}

\newcommand{\Fref}[1]{Figure~\ref{#1}}

\usepackage[pagebackref,breaklinks,colorlinks]{hyperref}

\usepackage[capitalize]{cleveref}
\crefname{section}{Sec.}{Secs.}
\Crefname{section}{Section}{Sections}
\Crefname{table}{Table}{Tables}
\crefname{table}{Tab.}{Tabs.}

\def\cvprPaperID{1981} 
\def\confName{CVPR}
\def\confYear{2022}

\begin{document}

\title{CSWin Transformer: A General Vision Transformer Backbone with Cross-Shaped Windows}

\author{
Xiaoyi Dong$^{1}$\thanks{Work done during an internship at Microsoft Research Asia.},  Jianmin Bao$^{2}$, Dongdong Chen$^{3}$, Weiming Zhang$^{1}$,\\ Nenghai Yu$^{1}$, Lu Yuan$^{3}$, Dong Chen$^{2}$, Baining Guo$^{2}$ \\
$^{1}$University of Science and Technology of China \\
$^{2}$Microsoft Research Asia
$^{3}$Microsoft Cloud + AI \\
{\tt\small\{dlight@mail., zhangwm@, ynh@\}.ustc.edu.cn } 
{\tt\small cddlyf@gmail.com }\\
{\tt\small\{jianbao, luyuan, doch, bainguo \}@microsoft.com } 
}

\maketitle

\begin{abstract}
   We present CSWin Transformer, an efficient and effective Transformer-based backbone for general-purpose vision tasks.
A challenging issue in Transformer design is that global self-attention is very expensive to compute whereas local self-attention often limits the field of interactions of each token. To address this issue, we develop the \emph{\textbf{C}ross-\textbf{S}haped \textbf{Win}dow} self-attention mechanism for computing self-attention in the horizontal and vertical stripes in \emph{parallel} that form a \emph{cross-shaped} window, with each stripe obtained by splitting the input feature into stripes of equal width. 
We provide a mathematical analysis of the effect of the stripe width and vary the stripe width for different layers of the Transformer network which achieves strong modeling capability while limiting the computation cost.
We also introduce \emph{Locally-enhanced Positional Encoding} (LePE), which handles the local positional information better than existing encoding schemes. LePE naturally supports arbitrary input resolutions, and is thus especially effective and friendly for downstream tasks. 
Incorporated with these designs and a hierarchical structure, CSWin Transformer demonstrates competitive performance on common vision tasks.
Specifically, it achieves \textbf{85.4\%} Top-1 accuracy on ImageNet-1K without any extra training data or label, \textbf{53.9} box AP and \textbf{46.4} mask AP on the COCO detection task, and \textbf{52.2} mIOU on the ADE20K semantic segmentation task, surpassing previous state-of-the-art Swin Transformer backbone by \textbf{+1.2}, \textbf{+2.0}, \textbf{+1.4}, and \textbf{+2.0} respectively under the similar FLOPs setting. By further pretraining on the larger dataset ImageNet-21K, we achieve \textbf{87.5\%} Top-1 accuracy on ImageNet-1K and high segmentation performance on ADE20K with \textbf{55.7} mIoU. 
\end{abstract}
\vspace{-4mm}
\section{Introduction}
Transformer-based architectures~\cite{dosovitskiy2020vit,touvron2020deit,liu2021swin,wu2021cvt} have recently achieved competitive performances compared to their CNN counterparts in various vision tasks. By leveraging the multi-head self-attention mechanism, these vision Transformers demonstrate a high capability in modeling the long-range dependencies, which is especially helpful for handling high-resolution inputs in downstream tasks, \eg, object detection and segmentation. Despite the success, the Transformer architecture with full-attention mechanism \cite{dosovitskiy2020vit}  is computationally inefficient.

\begin{figure*}[t]
\centering
\includegraphics[width=2\columnwidth]{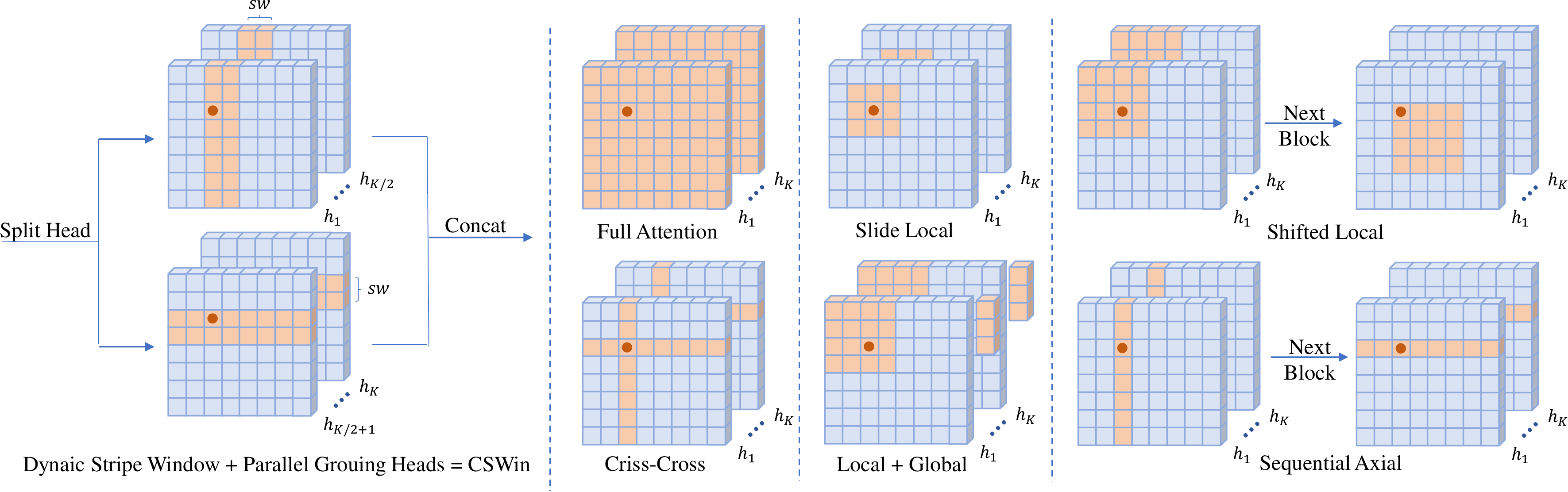} 
\vspace{-3mm}
\caption {Illustration of different self-attention mechanisms, our CSWin is fundamentally different from two aspects. First, we split multi-heads (\{$h_1,\dots,h_K$\}) into two groups and perform self-attention in horizontal and vertical stripes simultaneously. Second, we adjust the stripe width according to the depth network, which can achieve better trade-off between computation cost and capability }

\label{fig:attn_com}
\vspace{-5mm}
\end{figure*}

To improve the efficiency, one typical way is to limit the attention region of each token from full-attention to local/windowed attention~\cite{liu2021swin, vaswani2021scaling}. To bridge the connection between windows, researchers further proposed halo and shift operations to exchange information through nearby windows. However, the receptive field is enlarged quite slowly and it requires stacking a great number of blocks to achieve global self-attention. A sufficiently large receptive field is crucial to the performance especially for the downstream tasks(\eg, object detection and segmentation). Therefore it is important to achieve large receptive filed efficiently while keeping the computation cost low.

In this paper, we present the \emph{Cross-Shaped Window} (CSWin) self-attention, which is illustrated in \Fref{fig:attn_com} and compared with existing self-attention mechanisms. With CSWin self-attention, we perform the self-attention calculation in the horizontal and vertical stripes in parallel, with each stripe obtained by splitting the input feature into stripes of equal width. This stripe width is an important parameter of the cross-shaped window because it allows us to achieve strong modelling capability while limiting the computation cost. Specifically, we adjust the stripe width according to the depth of the network: small widths for shallow layers and larger widths for deep layers. A larger stripe width encourages a stronger connection between long-range elements and achieves better network capacity with a small increase in computation cost. We will provide a mathematical analysis of how the stripe width affects the modeling capability and computation cost.

\begin{figure*}[t]
\centering
\includegraphics[width=2\columnwidth]{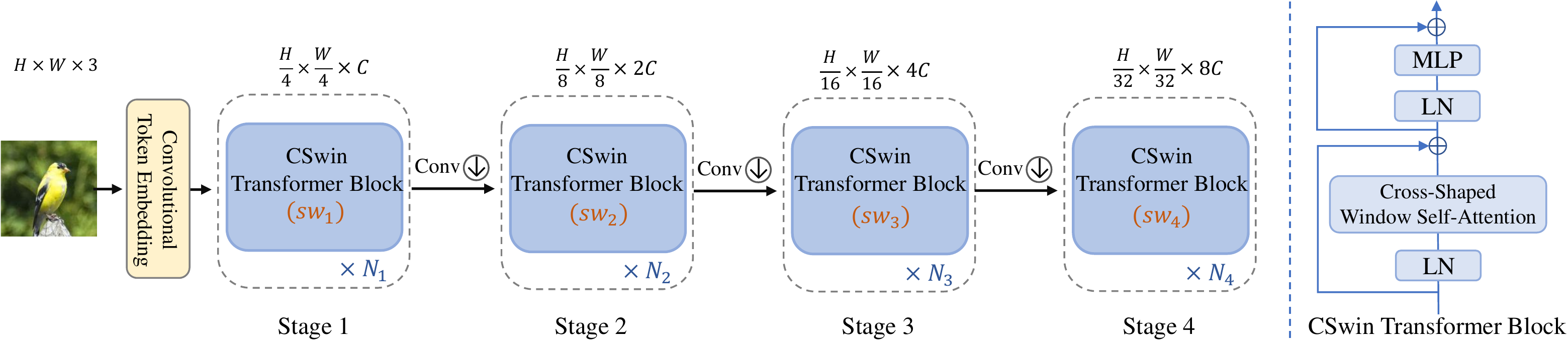} 

\caption{Left: the overall architecture of our proposed CSWin Transformer, Right: the illustration of  CSWin Transformer block.}
\label{fig:arch}
\vspace{-6mm}
\end{figure*}

It is worthwhile to note that with CSWin self-attention mechanism, the self-attention in horizontal and vertical stripes are calculated in parallel. We split the multi-heads into \textbf{parallel} groups and apply different self-attention operations onto different groups. This parallel strategy introduces no extra computation cost while enlarging the area for computing self-attention within each Transformer block. This strategy is fundamentally different from existing self-attention mechanisms~\cite{vaswani2017attention, liu2021swin, zhang2021mvit, ho2019axial} that apply the same attention operation across multi-heads((Figure~\ref{fig:attn_com} b,c,d,e), and perform different attention operations \textbf{sequentially}(Figure~\ref{fig:attn_com} c,e). We will show through ablation analysis that this difference makes CSWin self-attention much more effective for general vision tasks. 

Based on the CSWin self-attention mechanism, we follow the hierarchical design and propose a new vision Transformer architecture named ``CSWin Transformer'' for general-purpose vision tasks. This architecture provides significantly stronger modeling power while limiting computation cost. To further enhance this vision Transformer, we introduce an effective positional encoding, \emph{Locally-enhanced Positional Encoding} (LePE), which is especially effective and friendly for input varying downstream tasks such as object detection and segmentation. Compared with previous positional encoding methods~\cite{vaswani2017attention, shaw2018self, chu2021conditional}, our LePE imposes the positional information within each Transformer block and directly operates on the attention results instead of the attention calculation. The LePE makes CSWin Transformer more effective and friendly for the downstream tasks.

As a general vision Transformer backbone, the CSWin Transformer demonstrates  strong performance on image classification, object detection and semantic segmentation tasks. Under the similar FLOPs and model size, CSWin Transformer variants significantly outperforms previous state-of-the-art (SOTA) vision Transformers. For example, our base variant CSWin-B achieves \textbf{85.4\%} Top-1 accuracy on ImageNet-1K without any extra training data or label, \textbf{53.9} box AP and \textbf{46.4} mask AP on the COCO detection task, \textbf{51.7} mIOU on the ADE20K semantic segmentation task, surpassing previous state-of-the-art Swin Transformer counterpart by \textbf{+1.2}, \textbf{+2.0}, \textbf{1.4} and \textbf{+2.0} respectively. Under a smaller FLOPs setting, our tiny variant CSWin-T even shows larger performance gains, \textit{i.e.,},  \textbf{+1.4} point on ImageNet classification, \textbf{+3.0} box AP, \textbf{+2.0} mask AP on COCO detection and \textbf{+4.6} on ADE20K segmentation. Furthermore, when pretraining CSWin Transformer on the larger dataset ImageNet-21K, we achieve \textbf{87.5\%} Top-1 accuracy on ImageNet-1K and high segmentation performance on ADE20K with \textbf{55.7} mIoU.

\vspace{-1mm}
\section{Related Work}
\vspace{-1mm}
\noindent \textbf{Vision Transformers.} Convolutional neural networks (CNN) have dominated the computer vision field for many years and achieved tremendous successes \cite{krizhevsky2012imagenet, simonyan2014very, szegedy2015going,he2016deep,huang2017densely, chen2017dual, hu2018squeeze, tan2019efficientnet,howard2017mobilenets, sandler2018mobilenetv2, sun2019deep}. Recently, the pioneering work ViT ~\cite{dosovitskiy2020vit} demonstrates that pure Transformer-based architectures can also achieve very competitive results, indicating the potential of handling the vision tasks and natural language processing (NLP) tasks under a unified framework. Built upon the success of ViT,  many efforts have been devoted to designing better Transformer based architectures for various vision tasks, including low-level image processing \cite{chen2020pre,wan2021high}, image classification ~\cite{touvron2020deit, yuan2021tokens,xu2021coat,chu2021we,han2021tnt, wang2021pyramid,chu2021twins,wu2021cvt, chu2021twins, yuan2021incorporating, jiang2021token, touvron2021going, el2021training, he2021transreid},  object detection \cite{carion2020end,zhu2020deformable} and semantic segmentation \cite{wang2020end,zheng2020rethinking,strudel2021segmenter}. Rather than concentrating on one special task, some recent works \cite{wang2021pyramid,zhang2021mvit,liu2021swin} try to design a general vision Transformer backbone for general-purpose vision tasks. They all follow the hierarchical Transformer architecture but adopt different self-attention mechanisms. The main benefit of the hierarchical design is to utilize the multi-scale features and reduce the computation complexity by progressively decreasing the number of tokens.  In this paper,we propose a new hierarchical vision Transformer backbone by introducing cross-shaped window self-attention and locally-enhanced positional encoding.

\noindent \textbf{Efficient Self-attentions.} In the NLP field, many efficient attention mechanisms \cite{child2019generating,rae2019compressive,choromanski2020rethinking,katharopoulos2020Transformers,kitaev2020reformer,tay2020sparse,roy2021efficient, beltagy2020longformer} have been designed to improve the Transformer efficiency for handling long sequences. Since the image resolution is often very high in vision tasks, designing efficient self-attention mechanisms is also very crucial. However, many existing vision Transformers \cite{dosovitskiy2020vit,touvron2020deit,yuan2021tokens,wu2021cvt} still adopt the original full self-attention, whose computation complexity is quadratic to the image size. To reduce the complexity, the recent vision Transformers \cite{liu2021swin, vaswani2021scaling} adopt the local self-attention mechanism \cite{ramachandran2019stand} and its shifted/haloed version to add the interaction across different local windows.
Besides, axial self-attention\cite{ho2019axial} and criss-cross attention\cite{huang2020ccnet} propose calculating attention within stripe windows along horizontal or/and vertical axis. While the performance of axial attention is limited by its sequential mechanism and restricted window size, criss-cross attention is inefficient in practice due to its overlapped window design and ineffective due to its restricted window size. They are the most related works with our CSWin, which could be viewed as a much general and efficient format of these previous works.

\noindent \textbf{Positional Encoding.} Since self-attention is permutation-invariant and ignores the token positional information, positional encoding is widely used in Transformers to add such positional information back. Typical positional encoding mechanisms include absolute positional encoding (APE) \cite{vaswani2017attention}, relative positional encoding (RPE) ~\cite{shaw2018self,liu2021swin} and conditional positional encoding (CPE) \cite{chu2021conditional}. 
APE and RPE are often defined as the sinusoidal functions of a series of frequencies or the learnable parameters, which are designed for a specific input size and are not friendly to varying input resolutions.
CPE takes the feature as input and can generate the positional encoding for arbitrary input resolutions. Then the generated positional encoding will be added onto the input feature. Our LePE shares a similar spirit as CPE, but proposes to add the positional encoding as a parallel module to the self-attention operation and operates on projected \emph{values} in each Transformer block. This design decouples positional encoding from the self-attention calculation, and can enforce stronger local inductive bias.

\begin{figure*}[t]
\centering
\includegraphics[width=2\columnwidth]{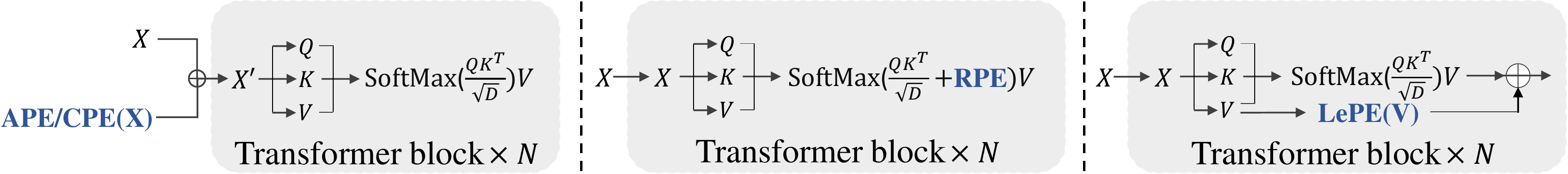} 
\vspace{-3mm}
\caption{Comparison among different positional encoding mechanisms: APE and CPE introduce the positional information before feeding into the Transformer blocks, while RPE and our LePE operate in each Transformer block. Different from RPE that adds the positional information into the attention calculation, our LePE operates directly upon $V$ and acts as a parallel module. {$^*$ Here we only draw the self-attention part to represent the Transformer block for simplicity.}}
\label{fig:pe_cmp}
\vspace{-4mm}
\end{figure*}

\vspace{-1mm}
\section{Method}
\vspace{-1mm}
\subsection{Overall Architecture}
\vspace{-1mm}

The overall architecture of CSWin Transformer is illustrated in Figure~\ref{fig:arch}. For an input image with size of $H \times W \times 3$, we follow \cite{wu2021cvt} and leverage  the overlapped convolutional token embedding ($7 \times 7$ convolution layer with stride 4) ) to obtain $\frac{H}{4} \times \frac{W}{4}$ patch tokens, and the dimension of each token is $C$.
To produce a hierarchical representation, the whole network consists of four stages. A convolution layer ($3 \times 3$, stride 2) is used between two adjacent stages to reduce the number of tokens and double the channel dimension. Therefore, the constructed feature maps have  $\frac{H}{2^{i+1}} \times \frac{W}{2^{i+1}}$ tokens for the $i^{th}$ stage, which is similar to traditional 
CNN backbones like VGG/ResNet. Each stage consists of $N_i$ sequential \emph{CSWin Transformer Blocks} and maintains the number of tokens. CSWin Transformer Block has the overall similar topology as the vanilla multi-head self-attention Transformer block with two differences: 1) It replaces the self-attention mechanism with our proposed Cross-Shaped Window Self-Attention; 2) In order to introduce the local inductive bias, LePE is added as a parallel module to the self-attention branch.

\subsection{Cross-Shaped Window Self-Attention}
Despite the strong long-range context modeling capability, the computation complexity of the original full self-attention mechanism is quadratic to feature map size. Therefore, it will suffer from huge computation cost for vision tasks that take high resolution feature maps as input, such as object detection and segmentation. To alleviate this issue, existing works~\cite{liu2021swin, vaswani2021scaling} suggest to perform self-attention in a local attention window and apply halo or shifted window to enlarge the receptive filed. However, the token within each Transformer block still has limited attention area and requires stacking more blocks to achieve global receptive filed. To enlarge the attention area and achieve global self-attention more efficiently, we present the cross-shaped window self-attention mechanism, which is achieved by performing self-attention in horizontal and vertical stripes in parallel that form a cross-shaped window.

\noindent \textbf{Horizontal and Vertical Stripes.}
According to the multi-head self-attention mechanism, the input feature $X\in\bm{R}^{(H\times W)\times C}$ will be first linearly projected to $K$ heads, and then each head will perform local self-attention within either the horizontal or vertical stripes.

For horizontal stripes self-attention, $X$ is evenly partitioned into non-overlapping horizontal stripes $[X^1,..,X^M]$ of equal width $sw$, and each of them contains $sw\times W$ tokens.
Here, $sw$ is the stripe width and can be adjusted to balance the learning capacity and computation complexity. Formally, suppose the projected queries, keys and values of the $k^{th}$ head all have dimension $d_k$, then the output of the horizontal stripes self-attention for $k^{th}$ head is defined as:

\begin{equation}
    \begin{aligned}
	&X = [X^1, X^2, \dots, X^M], \\
	&Y^i_k = \text{Attention}(X^iW^Q_k, X^iW^K_k, X^iW^V_k),\\
	&\mathrm{H\text{-}Attention}_k(X) = [Y^1_k, Y^2_k, \dots, Y^M_k]
    \end{aligned}
\end{equation}
Where $\text{where}~ X^i \in \bm{R}^{(sw\times W)\times C} $ and $M = H/sw$, $i = 1, \dots, M$. $W^Q_k \in \bm{R}^{C \times d_k}$, $W^K_k \in \bm{R}^{C \times d_k}$, $W^V_k \in \bm{R}^{C \times d_k}$ represent the projection  matrices of queries, keys and values for the $k^{th}$ head respectively, and $d_k$ is set as $C/K$.  The vertical stripes self-attention can be similarly derived, and its output for $k^{th}$ head is denoted as $\mathrm{V\text{-}Attention}_k(X)$.

Assuming natural images do not have directional bias, we equally split the $K$ heads into two parallel groups (each has $K/2$ heads, $K$ is often an even value). The first group of heads perform horizontal stripes self-attention  while the second group of heads perform vertical stripes self-attention. Finally the output of these two parallel groups will be concatenated back together.
\begin{small}
\vspace{-1mm}
\begin{equation}
    \begin{aligned}
	&\mathrm{CSWin\text{-}Attention}(X) = \mathrm{Concat}(\mathrm{head_1}, ..., \mathrm{head_K})W^O\\
	&\mathrm{head_k} = 
	\begin{cases}
		\mathrm{H\text{-}Attention}_k(X) & k = 1,\dots, K/2 \\
		\mathrm{V\text{-}Attention}_k(X) & k = K/2+1,\dots, K
	\end{cases}
\end{aligned}
\end{equation}
\vspace{-3mm}
\end{small}

Where $W^O\in \bm{R}^{C \times C}$ is the commonly used projection matrix that projects the self-attention results into the target output dimension (set as $C$ by default). As described above, one key insight in our self-attention mechanism design is splitting the multi-heads into different groups and applying different self-attention operations accordingly. In other words, \emph{the attention area of each token within one Transformer block is enlarged via multi-head grouping}. By contrast, existing self-attention mechanisms apply the same self-attention operations across different multi-heads. In the experiment parts, we will show that this design will bring better performance.

\noindent \textbf{Computation Complexity Analysis.} The computation complexity of CSWin self-attention is:
\begin{align}
\Omega (\text{CSWin}) = HWC*(4C + sw*H + sw*W)
\end{align}

For high-resolution inputs, considering $H,W$ will be larger than $C$ in the early stages and smaller than $C$ in the later stages, we choose small $sw$ for early stages and larger $sw$ for later stages. In other words,  \emph{adjusting $sw$ provides the flexibility to enlarge the attention area of each token in later stages in an efficient way}. Besides, to make the intermediate feature map size divisible by $sw$ for $224\times 224$ input, we empirically set $sw$ to $1,2,7,7$ for four stages by default.

\noindent \textbf{Locally-Enhanced Positional Encoding.} Since the self-attention operation is permutation-invariant, it will ignore the important positional information within the 2D image. To add such information back, different positional encoding mechanisms have been utilized in existing vision Transformers. In \Fref{fig:pe_cmp}, we show some typical positional encoding mechanisms and compare them with our proposed locally-enhanced positional encoding. In details, APE~\cite{vaswani2017attention} and CPE~\cite{chu2021conditional} add the positional information into the input token before feeding into the Transformer blocks, while RPE~\cite{shaw2018self} and our LePE incorporate the positional information within each Transformer block. But different from RPE that adds the positional information within the attention calculation (i.e., $\text{Softmax}(QK^T)$), we consider a more straightforward manner and impose the positional information upon the linearly projected \emph{values}. Meanwhile, we notice that RPE introduces bias in a per head manner, while our LePE is a per-channel bias, which may show more potential to serve as positional embeddings.

Mathematically, we denote the input sequence as $x=(x_1, \dots, x_n)$ of $n$ elements, and the output of the attention $z=(z_1, \dots, z_n)$ of the same length, where $x_i, z_i \in R^C$. Self-attention computation could be formulated as:

\begin{equation}
\vspace{-1mm}
    \label{eq:attention}
    z_i = \sum_{j=1}^n{\alpha_{ij}v_j}, \alpha_{ij}=exp(q_i^Tk_j/\sqrt{d})
\vspace{-1mm}
\end{equation}

where $q_i, k_i, v_i$ are the $queue, key$ and $value$ get by a linear transformation of the input $x_i$ and $d$ is the feature dimension. Then our Locally-Enhanced position encoding performs as a learnable per-element bias and Eq.\ref{eq:attention} could be formulated as:
\begin{equation}
\vspace{-1mm}
    z_i^k = \sum_{j=1}^n{(\alpha_{ij}^k + \beta_{ij}^k)v_j^k}
\vspace{-1mm}
\end{equation}
where $z_i^k$ represents the $k^{th}$ element of vector $z_i$. To make the LePE suitable to varying input size, we set a distance threshold to the LePE and set it to $0$ if the Chebyshev distance of token $i$ and $j$ is greater than a threshold $\tau$  ($\tau=3$ in the default setting).

\subsection{CSWin Transformer Block} 
Equipped with the above self-attention mechanism and positional embedding mechanism, CSWin Transformer block is formally defined as:
\begin{align}
    &{{\hat{{X}}}^{l}} = \text{CSWin-Attention}\left( {\text{LN}\left( {{{{X}}^{l - 1}}} \right)} \right) + {{X}}^{l - 1} ,\nonumber\\
    &{{{X}}^l} = \text{MLP}\left( {\text{LN}\left( {{{\hat{{X}}}^{l}}} \right)} \right) + {{\hat{{X}}}^{l}},
\end{align}
where ${{X}}^l$ denotes the output of $l$-th Transformer block or the precedent convolutional layer of each stage.

\begin{table}
\begin{center}
\small
\resizebox{\linewidth}{!}{
\setlength{\tabcolsep}{0.3mm}{
\vspace{-0.5mm}
\begin{tabular}{l@{\hspace{3pt}}|c@{\hspace{3pt}}|c|c|c|c|c}

\toprule
Models & \#Dim & \#Blocks& $sw$ & \#heads &\#Param. & FLOPs  \\
\midrule
		CSWin-T & 64  & 1,2,21,1 & 1,2,7,7 & 2,4,8,16 & 23M  & 4.3G \\
		CSWin-S & 64  & 2,4,32,2 & 1,2,7,7 & 2,4,8,16 & 35M  & 6.9G\\
		CSWin-B & 96  & 2,4,32,2 & 1,2,7,7 & 4,8,16,32 & 78M  & 15.0G \\
		CSWin-L & 144 & 2,4,32,2 & 1,2,7,7 & 6,12,24,48  & 173M & 31.5G \\
\bottomrule
\end{tabular}}}
\vspace{-3mm}
\caption{Detailed configurations of different variants of CSWin Transformer. The FLOPs are calculated with $224\times224$ input.}
\vspace{-10mm}
\label{tab:model_config}
\end{center}
\end{table}

\subsection{Architecture Variants}
For a fair comparison with other vision Transformers under similar settings, we build four different variants of CSWin Transformer as shown in \Tref{tab:model_config}: CSWin-T (Tiny), CSWin-S (Small), CSWin-B (Base), CSWin-L (Large). They are designed by changing the base channel dimension $C$ and the block number of each stage. In all these variants, the expansion ratio of each MLP is  set as $4$.  The head number of the four stages is set as $2,4,8,16$ in the first three variants and $6,12,24,48$ in the last variant respectively.

\vspace{-1mm}
\section{Experiments}
\vspace{-1mm}
\label{sec:experiment}
To show the effectiveness of CSWin Transformer as a general vision backbone, we conduct experiments on ImageNet-1K~\cite{deng2009imagenet} classification, COCO~\cite{lin2014microsoftcoco} object detection, and ADE20K~\cite{zhou2017scene} semantic segmentation. We also perform comprehensive ablation studies to analyze each component of CSWin Transformer. 
As most of the methods we compared did not report downstream inference speed, we use an extra section to report it for simplicity.

\begin{table}[t]
\renewcommand\arraystretch{.8}
\centering
\tablestyle{6pt}{1.05}

\resizebox{\linewidth}{!}{
\setlength{\tabcolsep}{.5mm}{
\begin{tabular}[t]{l|cccc|c}
\toprule
Method &  Image Size& \#Param. & FLOPs & Throughput& Top-1 \\
\midrule
Eff-B4~\cite{tan2019efficientnet}    &$380^2$  & 19M & 4.2G  & 349/s & \textbf{82.9} \\
Eff-B5~\cite{tan2019efficientnet}    &$456^2$  & 30M & 9.9G  & 169/s & \textbf{83.6} \\
Eff-B6~\cite{tan2019efficientnet}    &$528^2$  & 43M & 19.0G & 96/s  & 84.0 \\
\midrule
DeiT-S~\cite{touvron2020deit}        &$224^2$  & 22M & 4.6G  & 940/s & 79.8 \\
DeiT-B~\cite{touvron2020deit}        &$224^2$  & 87M & 17.5G & 292/s & 81.8 \\
DeiT-B~\cite{touvron2020deit}        &$384^2$  & 86M & 55.4G & 85/s  & 83.1 \\
\midrule
PVT-S~\cite{wang2021pyramid}         &$224^2$  & 25M & 3.8G  & 820/s & 79.8 \\
PVT-M~\cite{wang2021pyramid}         &$224^2$  & 44M & 6.7G  & 526/s & 81.2 \\
PVT-L~\cite{wang2021pyramid}         &$224^2$  & 61M & 9.8G  & 367/s & 81.7 \\
\midrule
T2T$_t$-14~\cite{yuan2021tokens}     &$224^2$  & 22M & 6.1G  & --  & 81.7 \\
T2T$_t$-19~\cite{yuan2021tokens}     &$224^2$  & 39M & 9.8G  & --  & 82.2 \\
T2T$_t$-24~\cite{yuan2021tokens}     &$224^2$  & 64M & 15.0G & --  & 82.6 \\
\midrule
CvT-13~\cite{wu2021cvt}              &$224^2$  & 20M & 4.5G  & --  & 81.6 \\
CvT-21~\cite{wu2021cvt}              &$224^2$  & 32M & 7.1G  & --  & 82.5 \\
CvT-21 ~\cite{wu2021cvt}             &$384^2$  & 32M & 24.9G & --  & 83.3 \\
\midrule
Swin-T~\cite{liu2021swin}            &$224^2$  & 29M & 4.5G  & 755/s & 81.3 \\
Swin-S~\cite{liu2021swin}            &$224^2$  & 50M & 8.7G  & 437/s & 83.0 \\
Swin-B~\cite{liu2021swin}            &$224^2$  & 88M & 15.4G & 278/s & 83.3 \\
Swin-B~\cite{liu2021swin}            &$384^2$  & 88M & 47.0G & 85/s  & 84.2 \\

\midrule
\rowcolor{Graylight} 
CSWin-T                              &$224^2$& 23M & 4.3G  & 701/s & 82.7 \\

\rowcolor{Graylight} 
CSWin-S                              &$224^2$& 35M & 6.9G  & 437/s & \textbf{83.6} \\
\rowcolor{Graylight} 
CSWin-B                              &$224^2$& 78M & 15.0G & 250/s & \textbf{84.2} \\
\rowcolor{Graylight} 
CSWin-B                              &$384^2$& 78M & 47.0G &     & \textbf{85.4} \\
\bottomrule
\end{tabular}}
}
\vspace{-3mm}
\caption{Comparison of different models on ImageNet-1K. } 
\vspace{-3mm}
\label{tab:imagenet}
\end{table}

\begin{table}[t]
\scriptsize 
\renewcommand\arraystretch{.85}
\centering
\resizebox{1.05\linewidth}{!}{
\setlength{\tabcolsep}{0.1mm}{
\begin{tabular}[t]{ccccc|ccccc}
\toprule
Method &   Param & Size & FLOPs & Top-1 & Method & Param & Size & FLOPs & Top-1\\
\midrule

R-101x3             & 388M & 384$^2$ & 204.6G & 84.4 & 
R-152x4             & 937M & 480$^2$ & 840.5G & 85.4 \\

\midrule

ViT-B/16           & 86M  & 384$^2$ & 55.4G  & 84.0 & 
ViT-L/16            & 307M & 384$^2$ & 190.7G & 85.2 \\

\midrule

\multirow{2}{*}{Swin-B } & \multirow{2}{*}{88M} & 224$^2$ & 15.4G & 85.2 &
\multirow{2}{*}{Swin-L } & \multirow{2}{*}{197M}& 224$^2$ & 34.5G & 86.3\\
& & 384$^2$ & 47.1G & 86.4 & & & 384$^2$ & 103.9G & 87.3\\

\midrule

\cellcolor{Graylight} & \cellcolor{Graylight} & \cellcolor{Graylight}224$^2$ & \cellcolor{Graylight}15.0G & \cellcolor{Graylight}85.9 &
\cellcolor{Graylight}& \cellcolor{Graylight} & \cellcolor{Graylight}224$^2$ & \cellcolor{Graylight}31.5G & \cellcolor{Graylight}86.5\\

\multirow{-2}{*}{\cellcolor{Graylight}CSWin-B }& \multirow{-2}{*}{\cellcolor{Graylight}78M} & \cellcolor{Graylight}384$^2$ & \cellcolor{Graylight} 47.0G & \cellcolor{Graylight}\textbf{87.0} & \multirow{-2}{*}{\cellcolor{Graylight}CSWin-L } &  \multirow{-2}{*}{\cellcolor{Graylight}173M} & \cellcolor{Graylight}384$^2$ & \cellcolor{Graylight}96.8G  & \cellcolor{Graylight}\textbf{87.5}\\

\bottomrule
\end{tabular}}
}
\vspace{-3mm}
\caption{ImageNet-1K fine-tuning results by pre-training on ImageNet-21K datasets. }
\vspace{-3mm}
\label{tab:imagenet22k}
\end{table}

\begin{table*}[ht]
\renewcommand\arraystretch{.85}
\begin{center}
\resizebox{\linewidth}{!}{
\setlength{\tabcolsep}{2mm}{
\begin{tabular}{l@{\hspace{3.2pt}}|c@{\hspace{3.2pt}}|c|c|c|c|c|c|c|c|c|c|c|c|c}
\toprule
\multirow{2}{*}{Backbone} & \#Params & FLOPs & \multicolumn{6}{c|}{Mask R-CNN 1x schedule} & \multicolumn{6}{c}{Mask R-CNN 3x + MS schedule}\\
 & (M) & (G) & $AP^b$ & $AP^b_{50}$ & $AP^b_{75}$ & $AP^m$ & $AP^m_{50}$ & $AP^m_{75}$ & $AP^b$ & $AP^b_{50}$ & $AP^b_{75}$ & $AP^m$ & $AP^m_{50}$ & $AP^m_{75}$ \\
\midrule
Res50~\cite{he2016deep}                & 44 & 260  & 38.0 & 58.6 & 41.4 & 34.4 & 55.1 & 36.7
& 41.0 & 61.7 & 44.9 & 37.1 & 58.4 & 40.1 \\
PVT-S~\cite{wang2021pyramid}           & 44 & 245  & 40.4 & 62.9 & 43.8 & 37.8 & 60.1 & 40.3
& 43.0 & 65.3 & 46.9 & 39.9 & 62.5 & 42.8 \\
ViL-S~\cite{zhang2021mvit}             & 45 & 218  & 44.9 & 67.1 & 49.3 & 41.0 & 64.2 & 44.1
& 47.1 & 68.7 & 51.5 & 42.7 & 65.9 & 46.2 \\
TwinsP-S~\cite{chu2021twins}           & 44 & 245  & 42.9 & 65.8 & 47.1 & 40.0 & 62.7 & 42.9
& 46.8 & 69.3 & 51.8 & 42.6 & 66.3 & 46.0  \\
Twins-S~\cite{chu2021twins}            & 44 & 228  & 43.4 & 66.0 & 47.3 & 40.3 & 63.2 & 43.4
& 46.8 & 69.2 & 51.2 & 42.6 & 66.3 & 45.8 \\
Swin-T~\cite{liu2021swin}              & 48 & 264  & 42.2 & 64.6 & 46.2 & 39.1 & 61.6 & 42.0
& 46.0 & 68.2 & 50.2 & 41.6 & 65.1 & 44.8 \\
\rowcolor{Graylight} 
CSWin-T                                & 42 & 279  & \textbf{46.7} & \textbf{68.6} & \textbf{51.3} & \textbf{42.2} & \textbf{65.6} & \textbf{45.4}
& \textbf{49.0} & \textbf{70.7} & \textbf{53.7} & \textbf{43.6} & \textbf{67.9} & \textbf{46.6}\\

\midrule
Res101~\cite{he2016deep}               & 63 & 336  & 40.4 & 61.1 & 44.2 & 36.4 & 57.7 & 38.8
& 42.8 & 63.2 & 47.1 & 38.5 & 60.1 & 41.3\\
X101-32~\cite{xie2017resx}             & 63 & 340  & 41.9 & 62.5 & 45.9 & 37.5 & 59.4 & 40.2
& 44.0 & 64.4 & 48.0 & 39.2 & 61.4 & 41.9 \\
PVT-M~\cite{wang2021pyramid}           & 64 & 302  & 42.0 & 64.4 & 45.6 & 39.0 & 61.6 & 42.1
& 44.2 & 66.0 & 48.2 & 40.5 & 63.1 & 43.5 \\
ViL-M~\cite{zhang2021mvit}             & 60 & 261  & 43.4 & ---- & ---- & 39.7 & ---- & ----
& 44.6 & 66.3 & 48.5 & 40.7 & 63.8 & 43.7 \\
TwinsP-B~\cite{chu2021twins}           & 64 & 302  & 44.6 & 66.7 & 48.9 & 40.9 & 63.8 & 44.2
& 47.9 & 70.1 & 52.5 & 43.2 & 67.2 & 46.3 \\
Twins-B~\cite{chu2021twins}            & 76 & 340  & 45.2 & 67.6 & 49.3 & 41.5 & 64.5 & 44.8
& 48.0 & 69.5 & 52.7 & 43.0 & 66.8 & 46.6  \\
Swin-S~\cite{liu2021swin}              & 69 & 354  & 44.8 & 66.6 & 48.9 & 40.9 & 63.4 & 44.2
& 48.5 & 70.2 & 53.5 & 43.3 & 67.3 & 46.6 \\
\rowcolor{Graylight} 
CSWin-S                                & 54 & 342  & \textbf{47.9} & \textbf{70.1} & \textbf{52.6} & \textbf{43.2} & \textbf{67.1} & \textbf{46.2}
& \textbf{50.0} & \textbf{71.3} & \textbf{54.7} & \textbf{44.5} & \textbf{68.4} & \textbf{47.7} \\

\midrule
X101-64~\cite{xie2017resx}             & 101 & 493  & 42.8 & 63.8 & 47.3 & 38.4 & 60.6 & 41.3
& 44.4 & 64.9 & 48.8 & 39.7 & 61.9 & 42.6 \\
PVT-L\cite{wang2021pyramid}            & 81  & 364  & 42.9 & 65.0 & 46.6 & 39.5 & 61.9 & 42.5
& 44.5 & 66.0 & 48.3 & 40.7 & 63.4 & 43.7 \\
ViL-B~\cite{zhang2021mvit}             & 76  & 365  & 45.1 & ---- & ---- & 41.0 & ---- & ----
& 45.7 & 67.2 & 49.9 & 41.3 & 64.4 & 44.5 \\
TwinsP-L~\cite{chu2021twins}           & 81  & 364  & 45.4 & ---- & ---- & 41.5 & ---- & ----
& ---- & ---- & ---- & ---- & ---- & ----\\
Twins-L~\cite{chu2021twins}            & 111 & 474  & 45.9 & ---- & ---- & 41.6 & ---- & ----
& ---- & ---- & ---- & ---- & ---- & ----\\
Swin-B~\cite{liu2021swin}              & 107 & 496  & 46.9 & ---- & ---- & 42.3 & ---- & ----
& 48.5 & 69.8 & 53.2 & 43.4 & 66.8 & 46.9\\
\rowcolor{Graylight} 
CSWin-B                                & 97  & 526  & \textbf{48.7} & \textbf{70.4} & \textbf{53.9} & \textbf{43.9} & \textbf{67.8} & \textbf{47.3}
& \textbf{50.8} & \textbf{72.1} & \textbf{55.8} & \textbf{44.9} & \textbf{69.1} & \textbf{48.3} \\
\bottomrule
\end{tabular}
}}
\end{center}
\vspace{-6mm}
\caption{Object detection and instance segmentation performance on the COCO val2017 with the Mask R-CNN framework. The FLOPs (G) are measured at resolution $800\times 1280$, and the models are pre-trained on the ImageNet-1K. ResNet/ResNeXt results are copied from ~\cite{wang2021pyramid}.}
\vspace{-4mm}
\label{tab:maskrcnn_comp_det}

\end{table*}

\subsection{ImageNet-1K Classification}
For fair comparison, we follow the training strategy in DeiT~\cite{touvron2020deit} as other baseline Transformer architectures \cite{wu2021cvt,liu2021swin}. Specifically, all our models are trained for 300 epochs with the input size of $224 \times 224$. We use the AdamW optimizer with weight decay of 0.05 for CSWin-T/S and 0.1 for CSWin-B. The default batch size and initial learning rate are set to 1024 and 0.001, and the cosine learning rate scheduler with 20 epochs linear warm-up is used. We apply increasing stochastic depth~\cite{huang2016deep} augmentation for CSWin-T, CSWin-S, and CSWin-B with the maximum rate as 0.1, 0.3, 0.5 respectively. 
When reporting the results of $384\times384$ input, we fine-tune the models for 30 epochs with the weight decay of $1e\text{-}8$, learning rate of $1e\text{-}5$, batch size of $512$.

In \Tref{tab:imagenet}, we compare our CSWin Transformer with state-of-the-art CNN and Transformer architectures. With the limitation of pages, we only compare with a few classical methods here and make a comprehensive comparison in the supplemental materials.

It shows that our CSWin Transformers outperform previous state-of-the-art vision Transformers by large margins. For example, CSWin-T achieves 82.7\% Top-1 accuracy with only 4.3G FLOPs, surpassing CvT-13, Swin-T and DeiT-S by 1.1\%, 1.4\% and 2.9\% respectively. And for the small and base model setting, our CSWin-S and CSWin-B also achieve the best performance. When finetuned on the $384\times384$ input, a similar trend is observed, which well demonstrates the powerful learning capacity of our CSWin Transformers. 

Compared with state-of-the-art CNNs, we find our CSWin Transformer is the only Transformer based architecture that achieves comparable or even better results than EfficientNet~\cite{tan2019efficientnet} under the small and base settings, while using less computation complexity . It is also worth noting that neural architecture search is used in EfficientNet but not in our CSWin Transformer design.

We further pre-train CSWin Transformer on ImageNet-21K dataset, which contains 14.2M images and 21K classes. Models are trained for 90 epochs with the input size of $224 \times 224$. We use the AdamW optimizer with weight decay of 0.1 for CSWin-B and 0.2 for CSWin-L, and the default batch size and initial learning rate are set to 2048 and 0.001. 
When fine-tuning on ImageNet-1K,  we train the models for 30 epochs with the weight decay of $1e\text{-}8$, learning rate of $1e\text{-}5$, batch size of $512$. The increasing stochastic depth~\cite{huang2016deep} augmentation for both CSWin-B and CSWin-L is set to 0.1. 

Table.\ref{tab:imagenet22k} reports the results of pre-training on ImageNet-21K. Compared to the results of CSWin-B pre-trained on ImageNet-1K, the large-scale data of ImageNet-21K brings a 1.6\%$\thicksim$1.7\% gain. CSWin-B and CSWin-L achieve 87.0\% and 87.5\% top-1 accuracy, surpassing previous methods.

\begin{table}[t]
\begin{center}
\resizebox{\linewidth}{!}{
\setlength{\tabcolsep}{1.mm}{
\begin{tabular}{l@{\hspace{3pt}}|c@{\hspace{3pt}}|c|c|c|c|c|c|c}
\toprule
\multirow{2}{*}{Backbone} & \#Params & FLOPs & \multicolumn{6}{c}{Cascade Mask R-CNN 3x +MS}\\
 & (M) & (G) & $AP^b$ & $AP^b_{50}$ & $AP^b_{75}$ & $AP^m$ & $AP^m_{50}$ & $AP^m_{75}$  \\
\midrule
Res50~\cite{he2016deep}        & 82 & 739  & 46.3 & 64.3 & 50.5 & 40.1 & 61.7 & 43.4 \\
Swin-T~\cite{liu2021swin}      & 86 & 745  & 50.5 & 69.3 & 54.9 & 43.7 & 66.6 & 47.1 \\
\rowcolor{Graylight} 
CSWin-T                        & 80 & 757  & \textbf{52.5} & \textbf{71.5} & \textbf{57.1} & \textbf{45.3} & \textbf{68.8} & \textbf{48.9} \\

\midrule

X101-32~\cite{xie2017resx}    & 101 & 819  & 48.1 & 66.5 & 52.4 & 41.6 & 63.9 & 45.2 \\
Swin-S~\cite{liu2021swin}     & 107 & 838  & 51.8 & 70.4 & 56.3 & 44.7 & 67.9 & 48.5 \\
\rowcolor{Graylight} 
CSWin-S                       & 92  & 820  & \textbf{53.7} & \textbf{72.2} & \textbf{58.4} & \textbf{46.4} & \textbf{69.6} & \textbf{50.6} \\

\midrule
X101-64~\cite{xie2017resx}    & 140 & 972  & 48.3 & 66.4 & 52.3 & 41.7 & 64.0 & 45.1 \\
Swin-B~\cite{liu2021swin}     & 145 & 982  & 51.9 & 70.9 & 56.5 & 45.0 & 68.4 & 48.7 \\
\rowcolor{Graylight} 
CSWin-B                       & 135 & 1004 & \textbf{53.9} & \textbf{72.6} & \textbf{58.5} & \textbf{46.4} & \textbf{70.0} & \textbf{50.4} \\
\bottomrule
\end{tabular}
}}
\vspace{-3mm}
\caption{Object detection and instance segmentation performance on the COCO val2017 with Cascade Mask R-CNN.}
\vspace{-3mm}
\label{tab:casaskrcnn_comp_det}
\end{center}
\vspace{-3mm}
\end{table}

\subsection{COCO Object Detection}
Next, we evaluate CSWin Transformer on the COCO objection detection task with the Mask R-CNN~\cite{he2017mask} and Cascade Mask R-CNN~\cite{cai2018cascade} framework respectively. Specifically, we pretrain the backbones on the ImageNet-1K dataset and follow the finetuning strategy used in Swin Transformer ~\cite{liu2021swin} on the COCO training set.

We compare CSWin Transformer with various backbones: previous CNN backbones ResNet~\cite{he2016deep}, ResNeXt(X)~\cite{xie2017aggregated},  and Transformer backbones  PVT~\cite{wang2021pyramid}, Twins~\cite{chu2021twins}, and Swin~\cite{liu2021swin}. Table~\ref{tab:maskrcnn_comp_det} reports the results of the Mask R-CNN framework with ``$1\times$'' (12 training epoch) and ``$3\times + \text{MS}$'' (36 training epoch with multi-scale training) schedule. It shows that our CSWin Transformer variants clearly outperforms all the CNN and Transformer counterparts. In details, our CSWin-T outperforms Swin-T by \textbf{+4.5} box AP, \textbf{+3.1} mask AP with the $1\times$ schedule and \textbf{+3.0} box AP, \textbf{+2.0} mask AP with the $3\times$ schedule respectively. We also achieve similar performance gain on small and base configuration.

Table~\ref{tab:casaskrcnn_comp_det} reports the results with the Cascade Mask R-CNN framework. Though Cascade Mask R-CNN is overall stronger than Mask R-CNN, we observe CSWin Transformers still surpass the counterparts by promising margins under different model configurations.

\begin{table}[t]

\centering
\resizebox{\linewidth}{!}{
\setlength{\tabcolsep}{.2mm}{
\begin{tabular}[t]{l|ccc|ccc}
\toprule
\multirow{2}{*}{Backbone} & \multicolumn{3}{c|}{Semantic FPN 80k} & \multicolumn{3}{c}{Upernet 160k}\\
 &   \#Param. & FLOPs & mIoU  &   \#Param. & FLOPs & SS/MS mIoU \\ 
\midrule
Res50~\cite{he2016deep}             & 28.5 & 183 & 36.7 & ----  & ---- & ----/---- \\
PVT-S~\cite{wang2021pyramid}        & 28.2 & 161 & 39.8 & ----  & ---- & ----/---- \\
TwinsP-S~\cite{chu2021twins}        & 28.4 & 162 & 44.3 & 54.6  & 919  & 46.2/47.5 \\
Twins-S~\cite{chu2021twins}         & 28.3 & 144 & 43.2 & 54.4  & 901  & 46.2/47.1 \\
Swin-T~\cite{liu2021swin}           & 31.9 & 182 & 41.5 & 59.9  & 945  & 44.5/45.8 \\
\rowcolor{Graylight} 
CSWin-T                      & 26.1 & 202 & \textbf{48.2} & 59.9  & 959  & \textbf{49.3}/\textbf{50.7} \\
\midrule
Res101~\cite{he2016deep}            & 47.5 & 260 & 38.8 & 86.0  & 1029 & ----/44.9 \\
PVT-M~\cite{wang2021pyramid}        & 48.0 & 219 & 41.6 & ----  & ---- & ----/---- \\
TwinsP-B~\cite{chu2021twins}        & 48.1 & 220 & 44.9 & 74.3  & 977  & 47.1/48.4 \\
Twins-B~\cite{chu2021twins}         & 60.4 & 261 & 45.3 & 88.5  & 1020 & 47.7/48.9 \\
Swin-S~\cite{liu2021swin}           & 53.2 & 274 & 45.2 & 81.3  & 1038 & 47.6/49.5 \\
\rowcolor{Graylight} 
CSWin-S                      & 38.5 & 271 & \textbf{49.2} & 64.6  & 1027 & \textbf{50.4}/\textbf{51.5} \\
\midrule
X101-64~\cite{xie2017resx}          & 86.4 & --- & 40.2 & ----  & ---- & ----/---- \\
PVT-L~\cite{wang2021pyramid}        & 65.1 & 283 & 42.1 & ----  & ---- & ----/---- \\
TwinsP-L~\cite{chu2021twins}        & 65.3 & 283 & 46.4 & 91.5  & 1041 & 48.6/49.8 \\
Twins-L~\cite{chu2021twins}         & 103.7& 404 & 46.7 & 133.0 & 1164 & 48.8/50.2 \\
Swin-B~\cite{liu2021swin}           & 91.2 & 422 & 46.0 & 121.0 & 1188 & 48.1/49.7 \\
\rowcolor{Graylight} 
CSWin-B                      & 81.2 & 464 & \textbf{49.9} & 109.2 & 1222 & \textbf{51.1}/\textbf{52.2} \\
\midrule
Swin-B$\dagger$~\cite{liu2021swin}           & ----  & ---- & ---- & 121.0 & 1841 & 50.0/51.7 \\
Swin-L$\dagger$~\cite{liu2021swin}           & ----  & ---- & ---- & 234.0 & 3230 & 52.1/53.5 \\
\rowcolor{Graylight} 
CSWin-B$\dagger$                       & ----  & ---- & ---- & 109.2 & 1941 & 51.8/52.6\\
\rowcolor{Graylight} 
CSWin-L$\dagger$                       & ----  & ---- & ---- & 207.7 & 2745 & \textbf{54.0}/\textbf{55.7}\\
\bottomrule
\end{tabular}}}
\vspace{-3mm}
\caption{Performance comparison of different backbones on the ADE20K segmentation task. Two different frameworks semantic FPN and Upernet are used. FLOPs are calculated with resolution $512\times2048$. ResNet/ResNeXt results and Swin FPN results are copied from ~\cite{wang2021pyramid} and ~\cite{chu2021twins} respectively. $\dagger$ means the model is pretrained on ImageNet-21K and finetuned with 640$\times$640 resolution.}
\vspace{-3mm}
\label{tab:segmentation}
\end{table}

\subsection{ADE20K Semantic Segmentation}
We further investigate the capability of CSWin Transformer for Semantic Segmentation on the ADE20K~\cite{zhou2017scene} dataset. Here we employ the semantic FPN~\cite{kirillov2019panoptic} and Upernet~\cite{xiao2018upernet} as the basic framework. For fair comparison, we follow previous works~\cite{wang2021pyramid,liu2021swin} and train Semantic FPN 80k iterations with batch size as 16, and Upernet 160k iterations with batch size as 16, more details are provided in the supplementary material. In \Tref{tab:segmentation}, we report the results of different methods in terms of mIoU and Multi-scale tested mIoU (MS mIoU). It can be seen that, our CSWin Transformers significantly outperform previous state-of-the-arts under different configurations. In details, CSWin-T, CSWin-S, CSWin-B achieve \textbf{+6.7}, \textbf{+4.0}, \textbf{+3.9} higher mIOU than the Swin counterparts with the Semantic FPN framework, and \textbf{+4.8}, \textbf{+2.8}, \textbf{+3.0} higher mIOU with the Upernet framework. Compared to the CNN counterparts, the performance gain is very promising and demonstrates the potential of vision Transformers again. When using the ImageNet-21K pre-trained model, our CSWin-L further achieves \textbf{55.7} mIoU  and surpasses the previous best model by +2.2 mIoU, while using less computation complexity.

\begin{table}[t]
\centering
\resizebox{1.02\linewidth}{!}{
\setlength{\tabcolsep}{0.8mm}{
\begin{tabular}[t]{l|cccc|cccc}
\toprule
\multirow{2}{*}{Model} &  \multicolumn{4}{c|}{Cascade Mask R-CNN on COCO } & \multicolumn{4}{c}{UperNet on ADE20K} \\
  & \#Param. & FLOPs & FPS& AP$^{b/m}$ & \#Param. & FLOPs& FPS & mIoU  \\
\midrule
Swin-T         & 86M & 745G & 15.3 & 50.5/43.7 
& 60M & 945G & 18.5 & 44.5 \\
\rowcolor{Graylight} 
CSWin-T        & 80M & 757G & 14.2 & \textbf{52.5}/\textbf{45.3} 
& 60M & 959G & 17.3 & \textbf{49.3} \\
\midrule
Swin-S        & 107M& 838G & 12.0 & 51.8/44.7 
& 81M & 1038G & 15.2 & 47.6 \\
\rowcolor{Graylight} 
CSWin-S       & 92M & 820G & 11.7 & \textbf{53.7}/\textbf{46.4} 
& 65M & 1027G & 15.6 & \textbf{50.4} \\
\midrule
Swin-B        & 145M & 982G & 11.2 & 51.9/45.0 
& 121M & 1188G & 9.92 & 48.1 \\
\rowcolor{Graylight} 
CSWin-B       & 135M & 1004G & 9.6 & \textbf{53.9}/\textbf{46.4} 
& 109M & 1222G & 9.08 & \textbf{51.1} \\
\bottomrule
\end{tabular}}}

\vspace{-3mm}
\caption{FPS comparison with Swin on downstream tasks.}
\label{tab:speed}
\vspace{-4mm}
\end{table}

\begin{table*}[t]
\centering
\resizebox{\linewidth}{!}{
\setlength{\tabcolsep}{1.mm}{
\begin{tabular}[t]{l|c|cccc|ccccc|cccc}
\toprule
\multirow{2}{*}{Model} & Attention &\multicolumn{4}{c|}{ImageNet} & \multicolumn{5}{c|}{COCO} & \multicolumn{4}{c}{ADE20K} \\
& Reigon & \#Param. & FLOPs & FPS  & Top1(\%)   & \#Param. & FLOPs & FPS& AP$^b$ & AP$^m$ 
& \#Param. & FLOPs & FPS &  mIoU(\%)  \\
\midrule

Axial    & H &   23M & 4.2G & 735 & 81.8 & 42M & 258G & 27.9 &  43.4  & 39.4  
& 26M & 186G & 50.3  & 42.6 \\
CSWin (fix sw=1)   & H  & 23M & 4.1G & 721 & 81.9 & 42M & 258G & 26.8 & 45.2 & 40.8 
& 26M & 179G & 49.1 & 47.5 \\
\midrule
Criss-Cross    & H*2-1  &   23M & 4.2G & 187 & 82.2 & 42M & 263G & 5.5 & 45.2  & 40.9 
& 26M & 186G & 17.6 & 47.4 \\
CSWin (fix sw=2)     & H*2          & 23M & 4.2G & 718 & 82.2 & 42M & 263G & 25.1 & 45.6 & 41.4  
& 26M & 186G & 47.2 & 47.6 \\

\midrule

CSWin (sw=1,2,7,7; Seq) & sw$\times$H  & 23M & 4.3G & 711 & 82.4 & 42M & 279G & 22.3 & 45.1 & 41.1  & 26M & 202G & 45.2 & 46.2 \\
\rowcolor{Graylight} 
CSWin (sw=1,2,7,7)   & sw$\times$H   & 23M & 4.3G & 701 & \textbf{82.7} & 42M & 279G & 21.1 & \textbf{46.7} & \textbf{42.2}  & 26M & 202G & 44.8 & \textbf{48.2} \\

\bottomrule
\end{tabular}}}

\vspace{-3mm}
\caption{Stripes-Based attention mechanism comparison. `Seq' means sequential multi-head attention like Axial-attention. `Attention Region' means the average number of tokens that each head calculates attention with.}
\label{tab:ccnet_comp}
\vspace{-3mm}
\end{table*}

\subsection{Inference Speed.}
Here we report the inference speed of our CSWin and Swin works. For downstream tasks, we report the FPS of Cascade Mask R-CNN for object detection on COCO and UperNet for semantic segmentation on ADE20K.
In most cases, the speed of our model is only slightly slower than Swin (less than 10\%), but our model outperforms Swin by large margins. For example, on COCO, CSWin-S are +1.9\% box AP and +1.7\% mask AP higher than Swin-S with similar inference speed(11.7 FPS vs. 12 FPS). Note that our CSWin-T performs better than Swin-B on box AP(+0.6\%), mask AP(+0.3\%) with much faster inference speed(14.2 FPS vs. 11.2 FPS), indicating our CSWin achieves better accuracy/FPS trade-offs.

\subsection{Ablation Study}
To better understand CSWin Transformers, we compare each key component with the previous works under a completely fair setting that we use the same architecture and hyper-parameter for the following experiments, and only vary one component for each ablation. For time consideration, we use Mask R-CNN with 1x schedule as the default setting for detection and instance segmentation evaluation, and Semantic FPN with 80k iterations and single-scale test for segmentation evaluation.

\noindent \textbf{Parallel Multi-Head Grouping.}
We first study the effectiveness of our novel ``Parallel Multi-Head Grouping'' strategy. Here we compare Axial-Attention~\cite{ho2019axial} and Criss-Cross-Attention~\cite{huang2020ccnet} \textbf{under the CSWin-T backbone}. ``Attention region'' is used as the computation cost metric for detailed comparison. To simplify, we assume the attention is calculated on a square input that $H=W$. 

In Table.\ref{tab:ccnet_comp}, we find that the ``parallel multi-head grouping'' is efficient and effective, especially for downstream tasks. When we replace the Parallel manner with Sequential, the performance of CSWin degrades on all tasks.
When comparing with previous methods under the similar attention region constrain, our $sw=1$ CSWin performs slightly better than Axial on ImageNet, while outperforming it by a large margin on downstream tasks. Our $sw=2$ CSWin performs slightly better than Criss-Cross Attention, while the speed of CSWin is $2\times \sim 5\times$ faster than it on different tasks, this further proves that our ``parallel'' design is much more efficient.

\begin{figure}[t]
\centering
\includegraphics[width=\columnwidth]{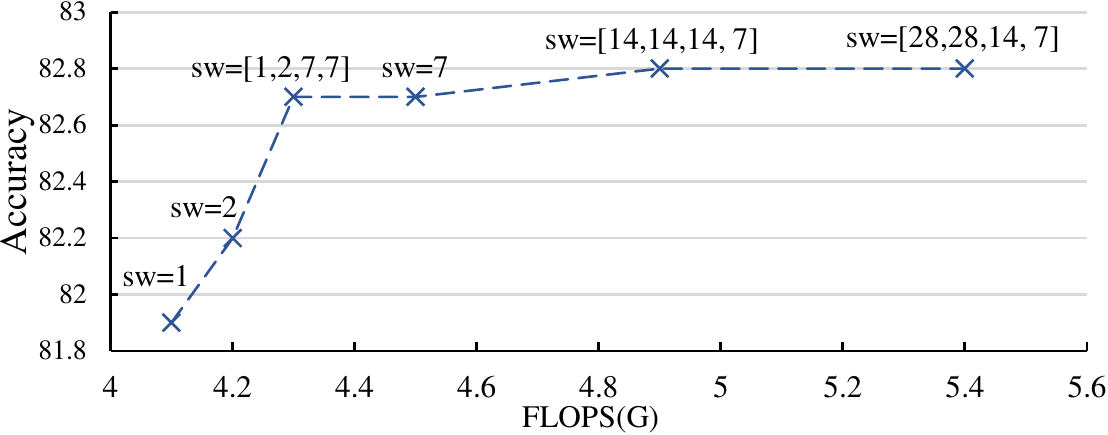} 
\vspace{-8mm}
\caption{Ablation on dynamic window size.}
\label{fig:sw}
\vspace{-4mm}
\end{figure}

\noindent \textbf{Dynamic Stripe Width .}
In Fig.\ref{fig:sw} we study the trade off between stripe width and accuracy. We find that with the increase of stripe width, the compution cost(FLOPS) increase, and the Top-1 classification accuracy improves greatly at the beginning and slows down when the width is large enough. Our default setting [1,2,7,7] achieves a good trade-off between accuracy and FLOPs.

\noindent \textbf{Attention Mechanism Comparison.} 
Following the above analysis on each component of CSWin self-attention, we further compare with existing self-attention mechanisms. As some of the methods need even layers in each stage, for a fair comparison, we use the \textbf{Swin-T~\cite{liu2021swin} as backbone and only change the self-attention mechanism}. In detail, we use $2, 2, 6, 2$ blocks for the four stages with the 96 base channel, non-overlapped token embedding~\cite{dosovitskiy2020vit}, and RPE~\cite{liu2021swin}. The results are reported in Table~\ref{tab:ablation_attn}. Obviously, our CSWin self-attention mechanism performs better than existing self-attention mechanisms across all the tasks.

\noindent \textbf{Positional Encoding Comparison.}
The proposed LePE is specially designed to enhance the local positional information on downstream tasks for various input resolutions. Here we use \textbf{CSWin-T as the backbone and only very the position encoding}. In
\Tref{tab:ablation_pe}, we compare our LePE with other recent positional encoding mechanisms(APE~\cite{dosovitskiy2020vit}, CPE~\cite{chu2021conditional}, and RPE~\cite{shaw2018self}) for image classification, object detection and image segmentation. Besides, we also test the variants without positional encoding (No PE) and CPE*, which is obtained by applying CPE before every Transformer block. According to the comparison results, we see that: 1) Positional encoding can bring performance gain by introducing the local inductive bias; 2) Though RPE achieves similar performance on the classification task with fixed input resolution, our LePE performs better (+1.2 box AP and +0.9 mask AP on COCO, +0.9 mIoU on ADE20K) on downstream tasks where the input resolution varies; 3) Compared to APE and CPE, our LePE also achieves better performance.

\begin{table}[t!]
\centering
\renewcommand\arraystretch{.8}
\resizebox{1.03\linewidth}{!}{
\setlength{\tabcolsep}{2.2mm}{
\begin{tabular}[t]{l|c|cc|c}
\toprule

 & \multicolumn{1}{c|}{ImageNet} & \multicolumn{2}{c|}{COCO} & \multicolumn{1}{c}{ADE20K} \\
 &  Top1(\%)     &  AP$^b$ & AP$^m$ & mIoU(\%)  \\
\midrule
Sliding window~\cite{ramachandran2019stand}    & 81.4 & --- & --- & ---- \\
Shifted window~\cite{liu2021swin}              & 81.3 & 42.2 & 39.1 & 41.5 \\
Spatially Sep~\cite{chu2021twins}              & 81.5 & 42.7 & 39.5 & 42.9 \\
Sequential Axial~\cite{ho2019axial}            & 81.5 & 40.4 & 37.6 & 39.8 \\
Criss-Cross~\cite{huang2020ccnet}              & 81.7 & 42.9 & 39.7 & 43.0 \\
\rowcolor{Graylight} 
Cross-shaped window                            & \textbf{82.2} & \textbf{43.4} & \textbf{40.2} & \textbf{43.4} \\
\bottomrule

\end{tabular}}}
\vspace{-3mm}
\caption{Comparison of different self-attention mechanisms.}
\vspace{-2mm}
\label{tab:ablation_attn}
\end{table}

\begin{table}[t!]
\renewcommand\arraystretch{.8}
\centering
\resizebox{1.03\linewidth}{!}{
\setlength{\tabcolsep}{3.8mm}{
\begin{tabular}[t]{l|c|cc|c}
\toprule

 & \multicolumn{1}{c|}{ImageNet} & \multicolumn{2}{c|}{COCO} & \multicolumn{1}{c}{ADE20K} \\
 &  Top1(\%)      &  AP$^b$ & AP$^m$ & mIoU(\%)  \\
\midrule
No PE                                & 82.5 & 44.8 & 41.1 & 47.0 \\
APE~\cite{dosovitskiy2020vit}        & 82.6 & 45.1 & 41.1 & 45.7 \\
CPE~\cite{chu2021conditional}        & 82.2 & 45.8 & 41.6 & 46.1 \\
CPE*~\cite{chu2021conditional}       & 82.4 & 45.4 & 41.3 & 46.6 \\
RPE~\cite{shaw2018self}              & \textbf{82.7} & 45.5 & 41.3 & 46.6 \\
\rowcolor{Graylight} 
LePE                                 & \textbf{82.7} & \textbf{46.7} & \textbf{42.2} & \textbf{48.2} \\

\bottomrule
\end{tabular}}}
\vspace{-3mm}
\caption{Comparison of different positional encoding mechanisms.}
\label{tab:ablation_pe}
\vspace{-4mm}
\end{table}

\section{Conclusion}
In this paper, we have presented a new Vision Transformer architecture named CSWin Transformer. The core design of CSWin Transformer is the CSWin Self-Attention, which performs self-attention in the horizontal and vertical stripes by splitting the multi-heads into \emph{parallel} groups. This multi-head grouping design can enlarge the attention area of each token within one Transformer block efficiently. On the other hand, the mathematical analysis also allows us to increase the stripe width along the network depth to further enlarge the attention area with subtle extra computation cost. We further introduce locally-enhanced positional encoding into CSWin Transformer for downstream tasks.  We achieved the state-of-the-art performance on various vision tasks under constrained computation complexity. We are  looking forward to applying it for more vision tasks.

{\small
\bibliographystyle{ieee_fullname}
\bibliography{egbib}
}

\newpage
\section*{Experiment Details}
In this section, we provide more detailed experimental settings about ImageNet and downstream tasks.

\noindent \textbf{ImageNet-1K Classification.}
For a fair comparison, we follow the training strategy in DeiT~\cite{touvron2020deit}. Specifically, all our models are trained for 300 epochs with the input size of $224 \times 224$. We use the AdamW optimizer with weight decay of 0.05 for CSWin-T/S and 0.1 for CSWin-B. The default batch size and initial learning rate are set to 2048 and $2e-3$ respectively, and the cosine learning rate scheduler with 20 epochs linear warm-up is used. We adopt most of the augmentation in ~\cite{touvron2020deit}, including RandAugment~\cite{cubuk2019randaugment} (rand-m9-mstd0.5-inc1) , Mixup~\cite{zhang2018mixup} $(prob=0.8)$, CutMix~\cite{yun2019cutmix} $(prob=1.0)$, Random Erasing ~\cite{zhong2017randomerase} $(prob=0.25)$ and  Exponential Moving Average ~\cite{polyak1992ema} $(ema$-$decay=0.99984)$, increasing stochastic depth~\cite{huang2016deep} ($prob=0.2, 0.4, 0.5$ for  CSWin-T, CSWin-S, and CSWin-B respectively).

When fine-tuning with $384\times384$ input, we follow the setting in ~\cite{liu2021swin} that fine-tune the models for 30 epochs with the weight decay of $1e\text{-}8$,  learning rate of $5e\text{-}6$, batch size of $256$. We notice that a large ratio of stochastic depth is beneficial for fine-tuning and keeping it the same as the training stage.

\noindent \textbf{COCO Object Detection and Instance Segmentation.}
We use two classical object detection frameworks: Mask R-CNN~\cite{he2017mask} and Cascade Mask R-CNN~\cite{cai2018cascade} based on the implementation from mmdetection~\cite{mmdetection}. For Mask R-CNN, we train it with ImageNet-1K pretrained model with two settings: $1\times$ schedule and $3\times$+MS schedule. For $1\times$ schedule, we train the model with single-scale input (image is resized to the shorter side of 800 pixels, while the longer side does not exceed 1333 pixels) for 12 epochs. We use AdamW~\cite{loshchilov2019adamw} optimizer with a learning rate of 0.0001, weight decay of 0.05 and batch size of 16. The learning rate declines at the 8 and 11 epoch with decay rate 0.1. The stochastic depth is also same as the ImageNet-1K setting that 0.1, 0.3, 0.5 for CSWin-T, CSWin-S, and CSWin-B respectively.
For $3\times$+MS schedule, we train the model with multi-scale input (image is resized to the shorter side between 480 and 800 while the longer side is no longer than 1333) for 36 epochs. The other settings are same as the $1\times$ except we decay the learning rate at epoch 27 and 33.
When it comes to Cascade Mask R-CNN, we use the same $3\times$+MS schedule as Mask R-CNN.

\noindent \textbf{ADE20K Semantic segmentation.}
Here we consider two semantic segmentation frameworks: UperNet~\cite{xiao2018upernet} and Semantic FPN~\cite{kirillov2019panoptic} based on the implementation from mmsegmentaion~\cite{mmseg2020}. 
For UperNet, we follow the setting in ~\cite{liu2021swin} and use AdamW~\cite{loshchilov2019adamw} optimizer with initial learning rate $6e^{-5}$, weight decay of 0.01 and batch size of 16 (8 GPUs with 2 images per GPU) for 160K iterations. The learning rate warmups with 1500 iterations at the beginning and decays with a linear decay strategy. 
We use the default augmentation setting in mmsegmentation including random horizontal flipping, random re-scaling (ratio range [0.5, 2.0]) and random photo-metric distortion. All the models are trained with input size $512\times512$. The stochastic depth is set to 0.2, 0.4, 0.6 for CSWin-T, CSWin-S, and CSWin-B respectively.
When it comes to testing, we report both single-scale test result and multi-scale test ([0.5, 0.75, 1.0, 1.25, 1.5, 1.75]$\times$ of that in training).

For Semantic FPN, we follow the setting in ~\cite{wang2021pyramid}. We use AdamW~\cite{loshchilov2019adamw} optimizer with initial learning rate $1e^{-4}$, weight decay of $1e^{-4}$ and batch size of 16 (4 GPUs with 4 images per GPU) for 80K iterations.

\section*{More Experimetns}
With the limitation of pages, we only compare with a few classical methods in our paper, here we make a comprehensive comparison with more current methods on ImageNet-1K. We find that our CSWin performs best in concurrent works.

\begin{table*}[t]
\scriptsize
\centering
\tablestyle{6pt}{1.05}
\begin{subtable}[t]{0.33\linewidth}
\resizebox{\linewidth}{!}{
\setlength{\tabcolsep}{1.99mm}{
\begin{tabular}[t]{l|cc|c}
\toprule
\multicolumn{4}{c}{\textbf{ImageNet-1K 224$^2$ trained models}} \\
Method &   \#Param. & FLOPs & Top-1 \\

\midrule
Reg-4G~\cite{radosavovic2020reg}           & 21M & 4.0G  & 80.0 \\
Eff-B4*~\cite{tan2019efficientnet}         & 19M & 4.2G  & \textbf{82.9} \\
\hline
DeiT-S~\cite{touvron2020deit}              & 22M & 4.6G & 79.8 \\
PVT-S~\cite{wang2021pyramid}               & 25M & 3.8G  & 79.8 \\
T2T-14~\cite{yuan2021tokens}               & 22M & 5.2G  & 81.5 \\
ViL-S~\cite{zhang2021mvit}                 & 25M & 4.9G  & 82.0 \\
TNT-S~\cite{han2021tnt}                    & 24M & 5.2G  & 81.3 \\
CViT-15~\cite{chen2021crossvit}            & 27M & 5.6G  & 81.0 \\
Visf-S~\cite{chen2021visformer}            & 40M & 4.9G  & 82.3 \\
LViT-S~\cite{li2021localvit}               & 22M & 4.6G  & 80.8 \\
CoaTL-S~\cite{xu2021coat}                  & 20M & 4.0G  & 81.9 \\
CPVT-S~\cite{chu2021conditional}           & 23M & 4.6G  & 81.5 \\
Swin-T~\cite{liu2021swin}                  & 29M & 4.5G  & 81.3 \\
CvT-13~\cite{wu2021cvt}                    & 20M & 4.5G  & 81.6 \\
\rowcolor{Graylight} 
CSWin-T                                    & 23M & 4.3G  & 82.7 \\
\midrule
\multicolumn{4}{c}{\textbf{ImageNet-1K 384$^2$ finetuned models}} \\
CvT-13 ~\cite{wu2021cvt}                   & 20M & 16.3G & 83.0 \\
T2T-14~\cite{yuan2021tokens}               & 22M & 17.1G & 83.3 \\
CViT$_{c}$-15~\cite{chen2021crossvit}      & 28M & 21.4G & 83.5 \\
\rowcolor{Graylight} 
CSWin-T                                    & 23M & 14.0G & \textbf{84.3} \\
\bottomrule
\end{tabular}}
}
\caption{Tiny Model}
\label{tab:tiny}
\end{subtable}
\hfill
\begin{subtable}[t]{0.33\linewidth}
\centering
\resizebox{0.985\linewidth}{!}{
\setlength{\tabcolsep}{2.01mm}{
\begin{tabular}[t]{l|cc|c}
\toprule
\multicolumn{4}{c}{\textbf{ImageNet-1K 224$^2$ trained models}} \\
Method &   \#Param. & FLOPs & Top-1 \\
\midrule
Reg-8G~\cite{radosavovic2020reg}           & 39M & 8.0G  & 81.7 \\
Eff-B5*~\cite{tan2019efficientnet}         & 30M & 9.9G  & \textbf{83.6} \\
\hline
PVT-M~\cite{wang2021pyramid}               & 44M & 6.7G  & 81.2 \\
PVT-L~\cite{wang2021pyramid}               & 61M & 9.8G  & 81.7 \\
T2T-19~\cite{yuan2021tokens}               & 39M & 8.9G  & 81.9 \\
T2T$_t$-19~\cite{yuan2021tokens}           & 39M & 9.8G  & 82.2 \\
ViL-M~\cite{zhang2021mvit}               & 40M & 8.7G  & 83.3 \\
MViT-B ~\cite{fan2021multiscale}           & 37M & 7.8G  & 83.0 \\
CViT-18~\cite{chen2021crossvit}            & 43M & 9.0G  & 82.5 \\
CViT$_{c}$-18~\cite{chen2021crossvit}      & 44M & 9.5G  & 82.8 \\
Twins-B~\cite{chu2021twins}                & 56M & 8.3G  & 83.2 \\
Swin-S~\cite{liu2021swin}                  & 50M & 8.7G  & 83.0 \\
CvT-21~\cite{wu2021cvt}                    & 32M & 7.1G  & 82.5 \\
\rowcolor{Graylight} 
CSWin-S                                    & 35M & 6.9G  & \textbf{83.6} \\
& & \\
\midrule
\multicolumn{4}{c}{\textbf{ImageNet-1K 384$^2$ finetuned models}} \\
CvT-21 ~\cite{wu2021cvt}                   & 32M & 24.9G & 83.3 \\
CViT$_{c}$-18~\cite{chen2021crossvit}      & 45M & 32.4G & 83.9 \\
\rowcolor{Graylight} 
CSWin-S                                    & 35M & 22.0G & \textbf{85.0} \\
& & \\

\bottomrule
\end{tabular}}}
\caption{Small Model}
\label{tab:small}
\end{subtable}
\hfill
\begin{subtable}[t]{0.33\linewidth}
\centering
\resizebox{0.985\linewidth}{!}{
\setlength{\tabcolsep}{1.98mm}{
\begin{tabular}[t]{l|cc|c}
\toprule
\multicolumn{4}{c}{\textbf{ImageNet-1K 224$^2$ trained models}} \\
Method &   \#Param. & FLOPs & Top-1 \\
\midrule
Reg-16G~\cite{radosavovic2020reg}          & 84M & 16.0G  & 82.9 \\
Eff-B6*~\cite{tan2019efficientnet}         & 43M & 19.0G  & 84.0 \\
\hline
DeiT-B~\cite{touvron2020deit}              & 87M & 17.5G  & 81.8 \\
PiT-B~\cite{heo2021pit}                    & 74M & 12.5G  & 82.0 \\
T2T-24~\cite{yuan2021tokens}               & 64M & 14.1G  & 82.3 \\
T2T$_t$-24~\cite{yuan2021tokens}           & 64M & 15.0G  & 82.6 \\
CPVT-B~\cite{chu2021conditional}           & 88M & 17.6G  & 82.3 \\
TNT-B~\cite{han2021tnt}                    & 66M & 14.1G  & 82.8 \\
ViL-B~\cite{zhang2021mvit}                 & 56M & 13.4G  & 83.2 \\
Twins-L~\cite{chu2021twins}                & 99M & 14.8G  & 83.7 \\
Swin-B~\cite{liu2021swin}                  & 88M & 15.4G  & 83.3 \\
\rowcolor{Graylight} 
CSWin-B                                    & 78M & 15.0G  & \textbf{84.2} \\
& & \\
& & \\
& & \\
\midrule
\multicolumn{4}{c}{\textbf{ImageNet-1K 384$^2$ finetuned models}} \\
ViT-B/16~\cite{dosovitskiy2020vit}         & 86M & 49.3G & 77.9 \\
DeiT-B~\cite{touvron2020deit}              & 86M & 55.4G & 83.1 \\
Swin-B~\cite{liu2021swin}                  & 88M & 47.0G & 84.2 \\
\rowcolor{Graylight} 
CSWin-B                                    & 78M & 47.0G  & \textbf{85.4} \\

\bottomrule
\end{tabular}}}
\caption{Base Model}
\label{tab:base}
\end{subtable}

\vspace{-1mm}
\caption{Comparison of different models on ImageNet-1K classification. * means the EfficientNet are trained with other input sizes. Here the models are grouped based on the computation complexity.} 
\vspace{-3mm}
\label{tab:imagenet}
\end{table*}

\end{document}